\documentclass[twocolumn]{fairmeta}
\usepackage[most]{tcolorbox}
\usepackage{tabularx}
\usepackage{makecell}
\usepackage{multirow}
\usepackage{amsmath,amssymb,amsfonts}
\usepackage{algorithmic}
\usepackage[percent]{overpic}
\usepackage{graphicx}
\usepackage{textcomp}
\usepackage{xcolor}
\usepackage{booktabs}
\usepackage{subcaption}
\usepackage{array}
\usepackage{caption}

\title{EM-Fall: Embodied mmWave Sensing for Day-and-Night Fall Detection on Humanoid Robots}

\author[1,2]{Yanshuo Lu}
\author[1,2]{Yuxuan Hu}
\author[1,3]{Shenghai Yuan}
\author[1,2]{Xinyu Zhou}
\author[1,2]{Kuangji Zuo}
\author[1,2]{Bofan Lyu}
\author[1,2]{XiChen Yuan}
\author[1,2]{Jianfei Yang}

\affiliation[1]{MARS Lab}
\affiliation[2]{NTU}
\affiliation[3]{IOT Lab}

\abstract{
Falls are one of the leading causes of injury and hospitalization among elderly individuals, making reliable fall awareness an essential capability for safety monitoring in residential environments. However, existing fall detection systems often rely on wearable devices or fixed sensing installations, which may suffer from low user compliance, limited spatial coverage, or degraded performance under occlusion and poor lighting conditions. In this work, we propose \textbf{EM-Fall}, an embodied fall detection framework deployed on a mobile humanoid robot. The system integrates millimeter-wave (mmWave) sensing with robotic mobility, allowing the robot to actively adjust its sensing viewpoint and maintain target observability across rooms and under occlusion. To address interference in complex residential environments, including pet motion and multipath artifacts, we design a human-centered perception pipeline combined with lightweight temporal modeling to capture motion evolution before, during, and after fall events. We evaluate the proposed system across eight real indoor environments with four participants and construct an in-home mmWave fall detection dataset. Experimental results show that the embodied mobile sensing paradigm improves monitoring continuity and maintains robust fall detection performance under diverse environmental conditions. The proposed framework provides a practical solution for robot-assisted safety monitoring in home environments.
}

\correspondence{Jianfei Yang at \email{jianfei.yang@ntu.edu.sg}}

\begin{document}

\maketitle

\section{Introduction}

With the rapid growth of the global aging population, enabling safe and independent living for elderly individuals has become an important societal challenge. Home service robots are increasingly envisioned as long-term companions that assist elderly people in daily residential environments. Among the various health and safety risks faced by older adults, falls are one of the leading causes of injury, hospitalization, and loss of independence. A large proportion of elderly individuals experience at least one fall each year, and delayed detection or response can lead to severe medical consequences. Therefore, equipping home robots with the capability to continuously monitor and promptly detect fall events is a critical requirement for robotic systems designed for elderly care~\citep{b1}. Achieving reliable fall monitoring in real residential environments has thus become an important research direction in home robotics~\citep{b30}.

\begin{figure}[htbp]
\centerline{\includegraphics[width=\columnwidth]{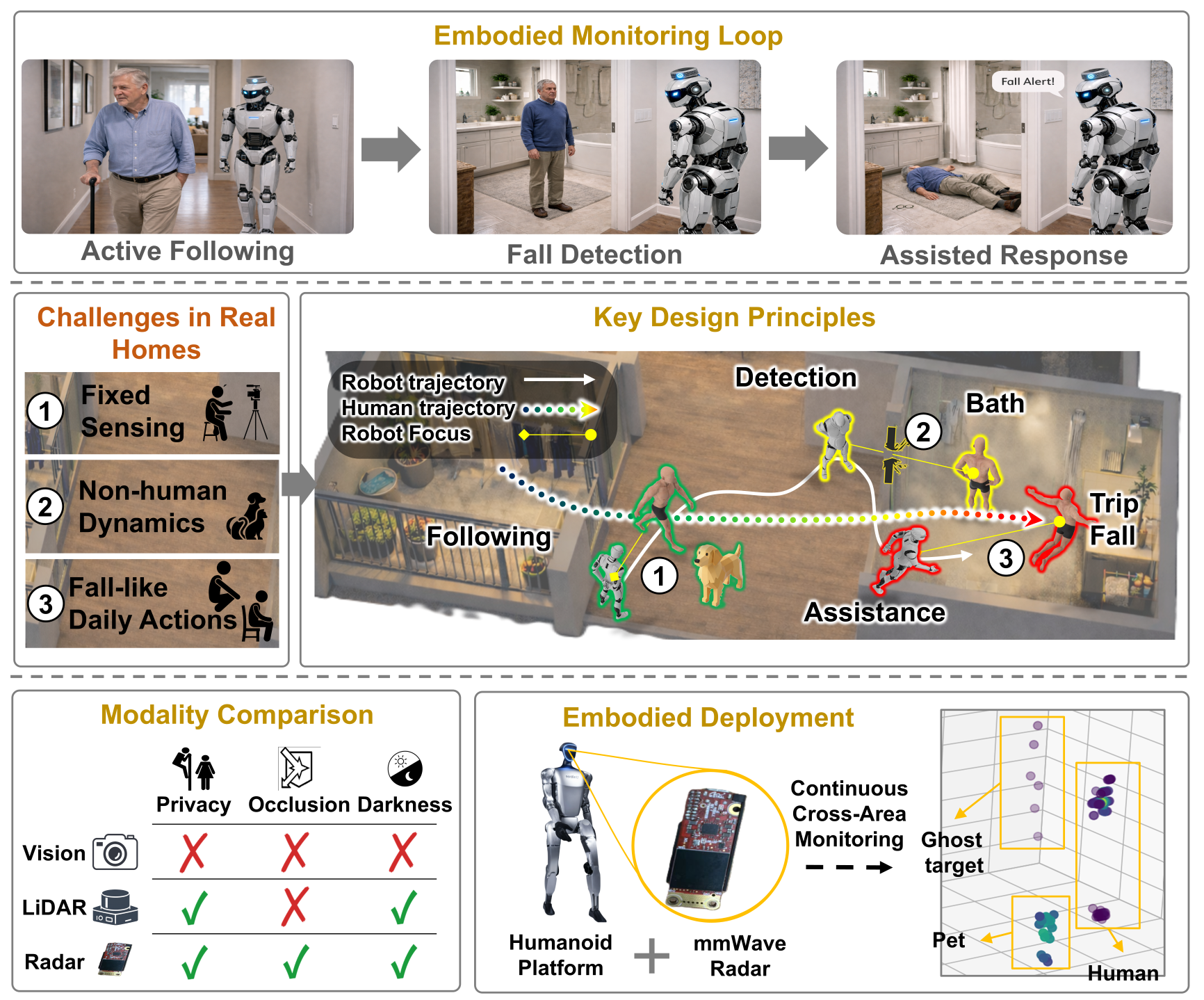}}
\caption{Overview of the proposed EM-Fall framework for embodied fall detection using mobile mmWave sensing. A humanoid robot performs continuous safety monitoring through user following, fall detection, and assisted response. By combining mmWave radar with robotic mobility, the system enables robust and continuous monitoring across rooms under occlusion, low-visibility, and privacy-sensitive conditions.}
\label{fig1}
\end{figure}

\begin{figure*}
\centerline{\includegraphics[width=\textwidth]{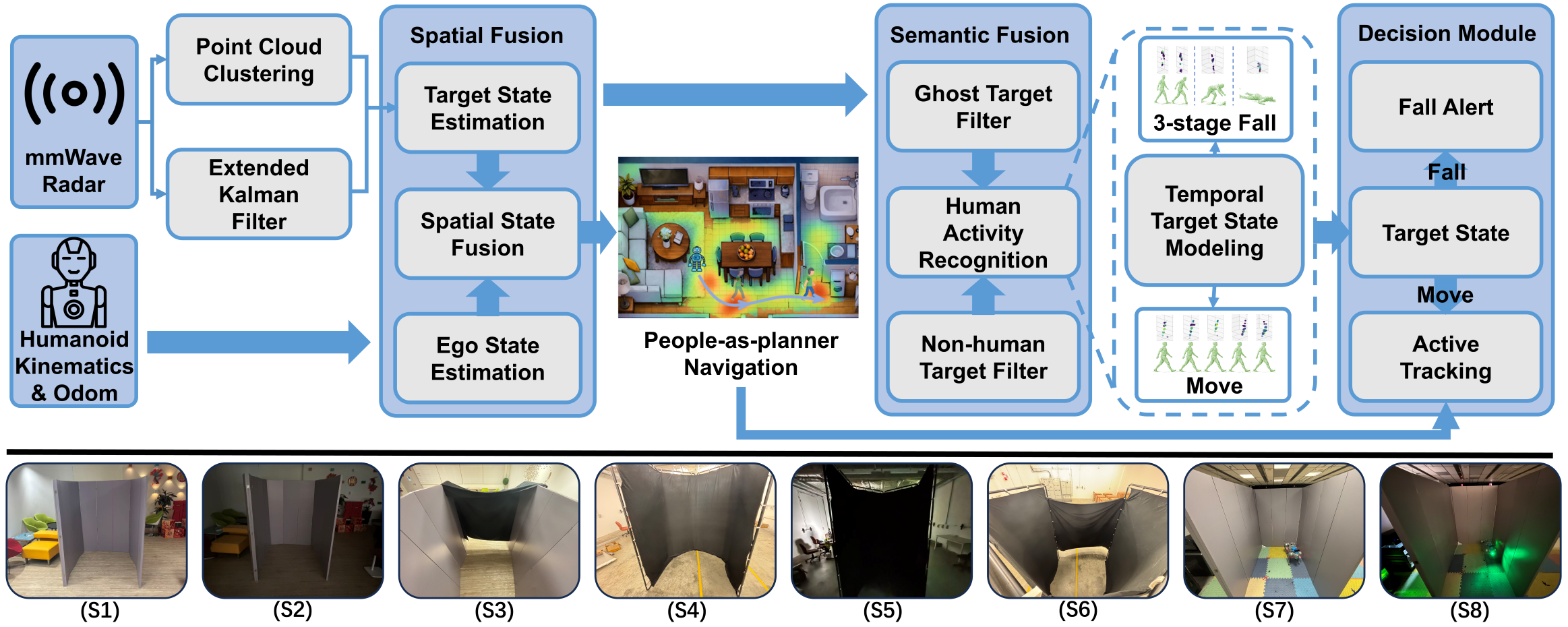}}
\caption{Overview of the EM-Fall framework. The upper panel illustrates the robotic perception pipeline: mmWave point clouds are clustered and tracked using EKF, fused with robot ego-motion for people-as-planner following, and processed through semantic fusion with a three-stage temporal module for fall detection. The lower panel presents the eight indoor evaluation scenarios (S1–S8) covering normal lighting, low-visibility, and occlusion conditions, including pet-like motion simulated by a quadruped robot. Dark scenes are exposure-adjusted for visualization.}
\label{fig2}
\end{figure*}
Existing fall detection solutions can generally be categorized into wearable sensing systems and contactless environmental sensing approaches. Wearable devices such as inertial sensors or smart watches~\citep{b32} provide direct motion measurements but require users to consistently wear the devices over long periods, which can be inconvenient and often leads to low user compliance in real-world deployment. Alternatively, contactless sensing methods have been widely explored. Vision-based systems~\citep{b33} can achieve good activity recognition performance under ideal conditions, yet their reliability significantly degrades under poor illumination or nighttime conditions and may raise privacy concerns in domestic environments. LiDAR-based sensing can capture geometric structure but may suffer from occlusion caused by obstacles such as furniture, curtains, or water mist in high-risk areas like bathrooms.

Millimeter-wave (mmWave) radar sensing provides a promising alternative for fall detection. Radar signals are insensitive to lighting conditions, preserve user privacy, and offer a certain level of penetration capability through obstacles~\citep{b34}. These advantages make radar particularly suitable for long-term monitoring in residential environments. However, most existing radar-based fall detection systems rely on fixed installations, which inevitably suffer from limited spatial coverage and occlusion-induced blind zones~\citep{b17}. When a fall occurs outside the sensing region or behind obstacles, the system may fail to observe the event. In contrast, mobile robots naturally provide an embodied sensing platform that can actively adjust their pose and viewpoint to maintain target observability and expand monitoring coverage across rooms. Integrating radar sensing with robotic mobility therefore provides a promising paradigm for continuous fall monitoring in real homes.

Despite these advantages, reliable fall detection in real residential environments remains challenging. In practice, many motion sources other than humans exist, including pet activities, swinging doors, moving furniture, and ghost reflections caused by multipath propagation. These factors introduce substantial environmental interference and can easily lead to false detections if not properly handled. Moreover, fall events exhibit strong temporal dynamics and may be difficult to distinguish from daily activities such as sitting or lying down without appropriate temporal modeling.

To address these challenges, we propose \textbf{EM-Fall}, an embodied mmWave fall detection framework deployed on a humanoid robotic platform. By integrating mmWave sensing with robotic mobility, the system maintains target observability and monitoring continuity across rooms and under occlusion. To improve robustness in complex residential environments, we design a hierarchical human-centered perception pipeline that performs non-human target filtering and temporal modeling of fall events. This enables reliable discrimination between true fall incidents and normal daily activities while suppressing interference caused by environmental motion sources.

We conduct comprehensive system-level validation across eight real indoor environments with four participants. Experimental results demonstrate that, compared with conventional fixed radar deployments, the embodied mobile sensing paradigm significantly improves monitoring continuity and robustness under environmental interference. The system effectively suppresses false alarms caused by non-human motion while maintaining reliable detection performance for real fall events.

The main contributions of this work are summarized as follows:

\begin{itemize}
\item We introduce an embodied mmWave fall detection paradigm that integrates radar sensing with robotic mobility, establishing a complete robotic pipeline from perception and event inference to decision and assistance. When deployed on a humanoid platform, the system improves cross-room monitoring continuity and alleviates coverage interruptions associated with static installations.

\item We develop a hierarchical human-centered perception pipeline that integrates target attribution, interference suppression, and temporal event modeling, enabling robust fall detection in complex residential environments with multiple environmental motion sources.

\item We conduct extensive real-world validation across eight indoor environments with four participants. The results demonstrate effective suppression of false alarms while maintaining reliable fall detection performance under diverse environmental conditions.
\end{itemize}

\section{Related works}

\subsection{Static Radar-based Fall Detection}

Radar-based fall detection has been widely studied as a privacy-preserving alternative to vision-based approaches. Most prior works adopt a static deployment paradigm, where radar sensors monitor predefined areas under a fixed sensing geometry. Early studies rely on spectral representations such as Doppler–time and range–velocity maps to capture rapid body descent dynamics. Subsequent works introduce multi-map fusion and deep models to improve generalization~\citep{b23}. With the development of MIMO mmWave radar, point cloud and 4D imaging representations have been adopted to preserve spatial structure and posture evolution~\citep{b9}. Temporal reasoning mechanisms are further incorporated to distinguish life-threatening falls from fall-like behaviors~\citep{b12}. Despite advances in representation and temporal modeling, these approaches remain constrained by fixed sensing geometry and predefined coverage areas.


To improve robustness in practical settings, recent studies explore mmWave radar for in-home fall detection. Multi-view fusion strategies~\citep{x1} and environment-invariant kinematic representations~\citep{x2} have been proposed to enhance generalization under environmental variation. However, many evaluations are conducted under relatively controlled assumptions, with limited consideration of additional motion sources such as pets and multipath-induced ghost reflections. These non-human dynamics significantly increase false detection rates and tracking instability. Moreover, although robustness to environmental variation is studied, sensing deployment largely remains static, and cross-area continuous monitoring is not explicitly addressed.

\subsection{Embodied Fall Detection Paradigms on Mobile Robots}

Mobile robotic platforms have been introduced to extend monitoring coverage in elderly care applications~\citep{c18,c20}. In most existing systems, fall detection is implemented using onboard cameras, while robot mobility primarily serves to reduce blind areas. Such approaches inherit the limitations of vision-based sensing, including sensitivity to illumination, occlusion, and privacy concerns. In addition, robot motion is typically driven by navigation objectives rather than perception-oriented sensing regulation. Radar-based fall detection under embodied mobile deployment has been rarely explored. Integrating mmWave radar into a humanoid robotic platform, therefore, represents a distinct sensing configuration that shifts fall detection from static installation to embodied mobile deployment, enabling continuous cross-area monitoring in realistic home environments.

\section{Methodology}

\subsection{Embodied Monitoring Architecture}
We propose an embodied mobile mmWave sensing paradigm for indoor caregiving that integrates perception, fall inference, and action generation into a unified robotic loop, as illustrated in Fig.~\ref{fig2}. 
The proposed system is organized into three tightly coupled layers: spatial state estimation for maintaining reliable target observability, human-centered perception and fall inference for robust activity understanding, and decision integration for generating robot actions.

The system first estimates human target states from mmWave radar point clouds and fuses them with robot ego-motion derived from kinematics and odometry to maintain spatial consistency under mobile deployment. 
The fused spatial representation enables people-as-planner tracking and lightweight local motion regulation for continuous following. To address occlusion, multipath effects, and non-human dynamics in real homes, a human-centered perception pipeline is introduced to suppress ghost targets, reject non-human entities, and model fall-related temporal state transitions. 
The structured state abstraction is then provided to a high-level decision module, which generates action policies for active tracking and assisted response, including fall alerting, thereby enabling continuous cross-area monitoring and event-driven intervention.
The decision module primarily serves as an integration interface between perception outputs and robot behavior generation, while the quantitative evaluation presented in this work focuses on the robustness of perception and fall inference.

\subsection{Spatial State Estimation and Active Regulation}
To maintain reliable spatial observability under mobile deployment, we first estimate stable human target trajectories from raw mmWave point clouds. Specifically, a clustering–EKF tracking pipeline~\citep{b45} is adopted to convert sparse and multipath-contaminated radar returns into temporally consistent target states.

Building upon the tracked human trajectory, we implement a people-as-planner following strategy~\citep{b46}. Instead of performing global replanning or predictive crowd modeling, the robot treats human motion as an implicit local guide and continuously regulates its motion to preserve monitoring continuity.

To compensate for robot ego-motion, the radar-estimated target pose is transformed into the robot coordinate frame as
\begin{equation}
{}^{R}\mathbf{T}_{O}(t)=
\begin{bmatrix}
{}^{R}\mathbf{R}_{S} & {}^{R}\mathbf{p}_{S}\\
\mathbf{0}^{\top} & 1
\end{bmatrix}
\begin{bmatrix}
{}^{S}\mathbf{R}_{O}(t) & {}^{S}\mathbf{p}_{O}(t)\\
\mathbf{0}^{\top} & 1
\end{bmatrix},
\label{eq:se3_compose}
\end{equation}
where ${}^{A}\mathbf{T}_{B}(t)\in SE(3)$ denotes the homogeneous transformation that maps coordinates from frame $\{B\}$ to frame $\{A\}$ at time $t$. Here, ${}^{R}\mathbf{R}_{S}\in SO(3)$ and ${}^{R}\mathbf{p}_{S}\in\mathbb{R}^{3}$ are the rotation matrix and translation vector of the radar frame $\{S\}$ with respect to the robot frame $\{R\}$, obtained from radar–robot extrinsic calibration. Moreover, ${}^{S}\mathbf{R}_{O}(t)\in SO(3)$ and ${}^{S}\mathbf{p}_{O}(t)\in\mathbb{R}^{3}$ denote the target orientation (approximated by the direction of the tracked planar velocity) and position expressed in $\{S\}$ at time $t$. The composition yields ${}^{R}\mathbf{T}_{O}(t)$, namely the target pose represented in the robot frame.

Let $p_R^t$ and $p_{H_L}^t$ denote the robot and human positions on the ground plane at time $t$, respectively. The subgoal located $d$ meters behind the leader along the robot--leader line is defined as
\begin{equation}
p_g^t = p_{H_L}^t - d \cdot \frac{p_{H_L}^t - p_R^t}{\left\|p_{H_L}^t - p_R^t\right\|}.
\label{eq:subgoal_behind_leader}
\end{equation}
We update the subgoal at a fixed frequency $f$ and maintain it over a short horizon to improve tracking stability. A lightweight local planner is then executed to reach the subgoal in real time. Let the leader velocity be $v_{H_L}^t \in \mathbb{R}^2$, and let the robot-leader distance be $\left\|p_{H_L}^t - p_R^t\right\|$. We directly modulate the maximum speed used by the local planner as
\begin{equation}
v_{\max}^{t}
=\min\!\left(
\bar{v},\ 
\left\lVert \mathbf{v}^{t}_{H_L} \right\rVert_{2}
+ k_{v}\big(\left\lVert \Delta \mathbf{p}^{t}\right\rVert_{2}-\rho\big)_{+}
\right),
\label{eq:vmax_modulation_linear}
\end{equation}
\begin{figure}[tbp]
\centerline{\includegraphics[width=\columnwidth]{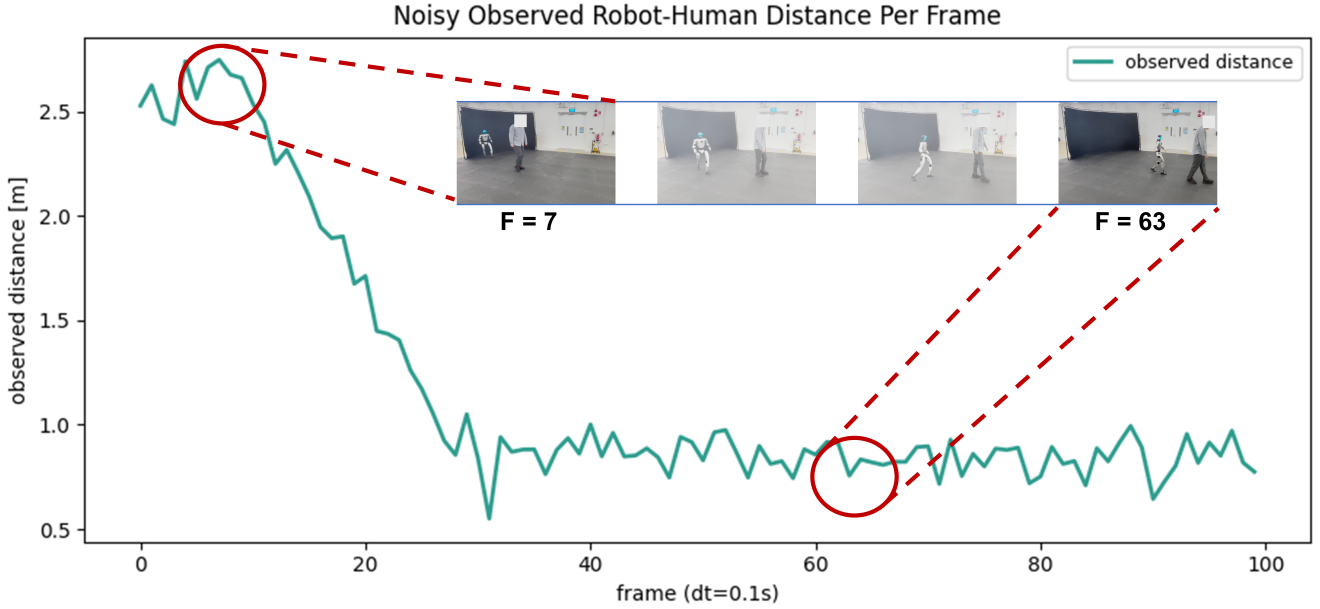}}
\caption{Illustration of people-as-planner active tracking. The observed robot–human distance is updated every 0.1\,s, while subgoals are refreshed over a 0.5\,s horizon to maintain stable following under noisy observations.}
\label{fig3}
\end{figure}
where $\bar{v}>0$ is a preset speed cap and typically $\bar{v}\ge \left\lVert \mathbf{v}^{t}_{L} \right\rVert_{2}$, $\mathbf{v}^{t}_{L}$ denotes the leader velocity on the ground plane, $\Delta \mathbf{p}^{t}$ is the robot--leader displacement on the ground plane, $\rho$ is the capture distance threshold, and $k_v>0$ is a linear gain that converts the excess distance into an additional speed allowance. The operator $(\cdot)_{+}$ is defined as
\begin{equation}
(x)_{+}\triangleq \max(x,0),
\end{equation}
so that the extra term is activated only when $\left\lVert \Delta \mathbf{p}^{t}\right\rVert_{2}>\rho$.
This design enables the mobile platform to perform responsive indoor following and continuous monitoring. The tracking performance is illustrated in Fig.~\ref{fig3}.

\subsection{Human-Centered Perception and Fall Inference}
Given the tracked target states, we process them with a hierarchical human-centered perception pipeline. Specifically, we refine raw target observations through a cascaded filtering procedure. A non-human target filter is first applied to remove returns associated with pets and other moving objects, followed by a ghost target filter to suppress spurious tracks caused by multipath propagation. For the remaining human targets, we perform human activity recognition using an MLP-based temporal state model that learns discriminative representations from continuous target state sequences and classifies the underlying activity. By explicitly modeling motion semantics and the state transitions surrounding fall events, the proposed cascade supports online fall recognition with improved robustness to measurement noise, background clutter, and multipath interference.

The non-human target filter aims to separate human targets from non-human entities. Since mmWave radars typically produce point clouds that mainly correspond to moving or dynamically changing object components~\citep{b24}, we aggregate the most recent frames into a stacked point cloud to obtain a more stable observation. Based on the aggregated data, we compute two geometric cues, namely the estimated target height and the number of associated points. A threshold-based rule is then used to reject non-human targets and retain human candidates according to these cues. To further improve robustness against frame-level fluctuations, we adopt a voting scheme over the temporal window to determine the final classification outcome.

The ghost target filter aims to remove ghost targets from the set of detected tracks. Although TI’s target tracking pipeline provides partial suppression of spurious tracks~\citep{b45}, our empirical evaluation indicates that ghost targets are still frequently produced in cluttered indoor environments. To further mitigate this issue, we introduce an additional state machine-based filter that operates on the tracked outputs. After applying the TI tracking method, each point in the point cloud is represented as a seven-dimensional feature vector:
\begingroup
\begin{equation}
\small
\mathbf{p}_i =
\begin{bmatrix}
x_i ,& y_i ,& z_i ,& v_i, & \mathrm{SNR}_i, & \mathrm{noise}_i, & \mathrm{id}_i
\end{bmatrix}
\in \mathbb{R}^{7},
\end{equation}
\endgroup
where $x_i$, $y_i$, and $z_i$ denote the three-dimensional Cartesian coordinates of the $i$-th point, $v_i$ is the corresponding Doppler velocity, $\mathrm{SNR}_i$ and $\mathrm{noise}_i$ are signal quality indicators, and $\mathrm{id}_i$ is the cluster identifier assigned by the TI tracking module. We then apply a threshold based screening rule using the per-cluster point count and aggregate $\mathrm{SNR}_i$ to reject low-confidence clusters and suppress spurious targets. Only clusters that satisfy these criteria are retained for subsequent state updates.

To enforce temporal consistency and suppress ghost targets, we associate each cluster $i$ with a discrete confidence state $c_i[t]\in\mathbb{Z}_{\ge 0}$, which is updated at every point cloud frame.
Let $\mathcal{I}_t$ denote the index set of clusters that are successfully observed (i.e., associated with valid measurements) at time step $t$.
We further introduce two thresholds, where $s_1$ specifies the confirmation level and $s_2$ controls the saturation margin of highly reliable tracks.
The state update rule is designed to (i) reward persistent observations by increasing $c_i[t]$ before confirmation, (ii) clamp the confidence to a maximum of $s_1+s_2$ once confirmed, and (iii) penalize missed detections via a one-step decay with a non-negativity constraint, as formalized in Equation 
\eqref{eq:confidence_update_cases},
\begingroup
\begin{equation}\footnotesize
\label{eq:confidence_update_cases}
c_i[t{+}1]=
\begin{cases}
c_i[t] + 1, & \bigl(i\in\mathcal{I}_t\bigr)\ \wedge\ \bigl(c_i[t] < s_1-1\bigr).\\[2pt]
s_1 + s_2,  & \bigl(i\in\mathcal{I}_t\bigr)\ \wedge\ \bigl(c_i[t] \ge s_1-1\bigr).\\[2pt]
\max\!\bigl(0,\, c_i[t]-1\bigr), & i\notin\mathcal{I}_t.
\end{cases}
\end{equation}
\endgroup
When $c_i[t]$ gradually increases and reaches $s_1$, the corresponding cluster is identified as a valid target, and an additional increment $s_2$ enhances its confidence level. If a cluster is not detected in subsequent frames, its state value decreases by one per frame until reaching zero, where it is considered a ghost target. This update mechanism suppresses transient false alarms and provides a reliable target tracking capability.

The filtered human target sequences are then used as inputs to the fall detection module. For each tracked target, we compute six lightweight geometric descriptors from the current point cloud together with a short temporal history. Specifically, the per-frame feature vector is defined as
\begin{equation}
\mathbf{f}^{t}=\big[h_{\mathrm{avg}}^{t},\,\Delta h^{t},\,\Delta z^{t},\,w_{\mathrm{xy}}^{t},\,n_{\mathrm{pt}}^{t},\,\ell_{z}^{t}\big]^{\top}.
\end{equation}
Here, $h_{\mathrm{avg}}^{t}$ denotes the recent average height of the target, $\Delta h^{t}$ denotes the height change over the temporal window, $\Delta z^{t}$ denotes the vertical displacement of the target center, $w_{\mathrm{xy}}^{t}$ denotes the planar spatial extent on the ground plane, $n_{\mathrm{pt}}^{t}$ denotes the normalized point count as a proxy of point cloud density or size, and $\ell_{z}^{t}$ denotes the vertical extent of the target. Together, they form a compact yet informative representation of spatiotemporal dynamics that supports robust fall-related state inference in the presence of clutter and partial observations.
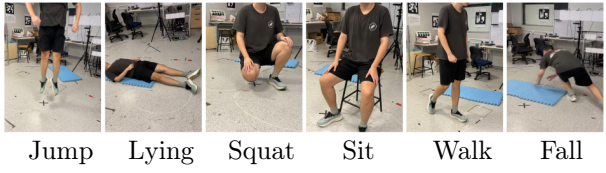
\begin{figure}[htbp]
  \centering
  \begin{overpic}[width=\columnwidth]{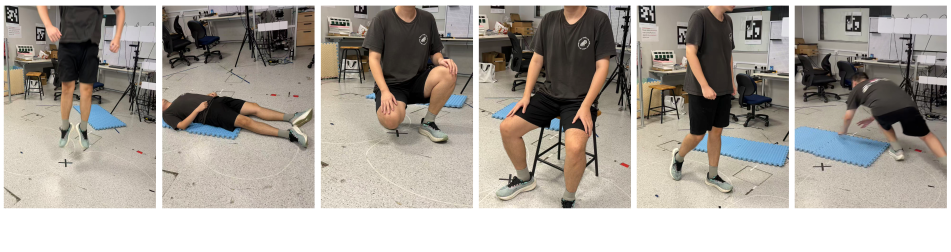}
    \put(4.2,0){Jump}
    \put(20.8,0){Lying}
    \put(37.4,0){Squat}
    \put(56.4,0){Sit}
    \put(71.4,0){Walk}
    \put(89.2,0){Fall}
  \end{overpic}
\caption{Six representative activities used in the evaluation: jump, lying, squat, sit, walk, and fall.}
\label{fig5}
\end{figure}

The six-dimensional feature tensors extracted from the point cloud are fed into a moment-based drop detector. Recent progress in neural networks has shown a strong capability in learning discriminative latent representations from noisy sensory measurements. We therefore adopt a three-layer multilayer perceptron (MLP) as the drop detector to predict whether a drop event occurs at the current time step.
Although temporal architectures are widely used for sequential modeling, their recursive computation can introduce non-negligible latency. Moreover, under limited training data, overly complex temporal models may exhibit reduced cross-scenario generalization~\citep{b28,r3}. During training, feature vectors corresponding to the drop phase in fall sequences are labeled as the positive class, whereas all feature vectors from non-fall sequences are labeled as the negative class. Given the pronounced class imbalance, we employ a weighted binary cross-entropy loss function to train the MLP drop detector, 
\begin{equation}
\footnotesize
\mathcal{L}_{\mathrm{wBCE}}
= -\frac{1}{N}\sum_{i=1}^{N}\Big( w_{+}\, y_i \log(\hat{y}_i) + w_{-}\,(1-y_i)\log(1-\hat{y}_i) \Big),
\label{eq:wbce}
\end{equation}
where $N$ is the number of training samples, $y_i\in\{0,1\}$ and $\hat{y}_i\in(0,1)$ denote the ground truth label and predicted probability for sample $i$, respectively, and $w_{+}>0$ and $w_{-}>0$ are class weights used to compensate for the imbalance.

If a drop action is detected at the current time step, the module waits for a short verification horizon, and then it computes the following post-event and pre-event metrics. First, it measures the maximum body width observed after the drop event,
\begin{equation}
W_{\max} \triangleq \max_{t=1,\ldots,\mathcal{T}_{\mathrm{post}}} W(t), 
\end{equation}
where $\mathcal{T}_{\mathrm{post}}$ denotes the corresponding post event evaluation window. Then it computes the proportion of post drop frames whose average height falls below a predefined threshold,
\begin{equation}
\label{eq:Rh}
R_h \triangleq \frac{1}{|\mathcal{T}_{\mathrm{post}}|}
\sum_{t\in\mathcal{T}_{\mathrm{post}}}
\mathbb{I}\!\left(\bar{h}(t) < h_{\mathrm{th}}\right),
\end{equation}
where $\mathbb{I}(\cdot)$ denotes the indicator function. In addition, the module evaluates the average delta height within the $k$ frames preceding the fall,
\begin{equation}
\small
\label{eq:Havg}
H_{\mathrm{avg}} \triangleq \frac{1}{|\mathcal{T}_{\mathrm{pre}}|}
\sum_{t\in\mathcal{T}_{\mathrm{pre}}} \Delta h(t), 
\mathcal{T}_{\mathrm{pre}} \triangleq \{t_f-k,\ldots,t_f-1\}.
\end{equation}
A threshold based decision rule is applied to fuse the three pooled features and infer the final fall label, 
\begin{equation}
z_{1}(t) = \mathbb{I}\!\left(H_{\mathrm{avg}}(t) < -\tau_{\Delta h}\right),
\label{eq:three_stage_rule_a}
\end{equation}
\begin{equation}
z_{2}(t) = \mathbb{I}\!\left(\bigl(W_{\max}(t) > \tau_{w}\bigr)\ \land\ \bigl(R_{h}(t) > \tau_{r}\bigr)\right),
\label{eq:three_stage_rule_b}
\end{equation}
\begin{equation}
\hat{y}(t) = \mathbb{I}\!\left(z_{1}(t)\ \land\ z_{2}(t)\ \land\ \mathcal{D}(t)\right),
\label{eq:three_stage_rule_d}
\end{equation}
where $\mathbb{I}(\cdot)$ denotes the indicator function, $\tau_{\Delta h}$, $\tau_{w}$, and $\tau_{r}$ are predefined thresholds, and $\mathcal{D}(t)\in\{0,1\}$ is the binary drop event indicator at time $t$. By jointly modeling characteristic patterns preceding the fall and post-fall signatures, this rule based fusion improves robustness and yields higher fall detection accuracy than approaches that rely on a single stage or a single cue.

\begin{figure*}[hbt]
  \centering
  \begin{overpic}[width=\textwidth]{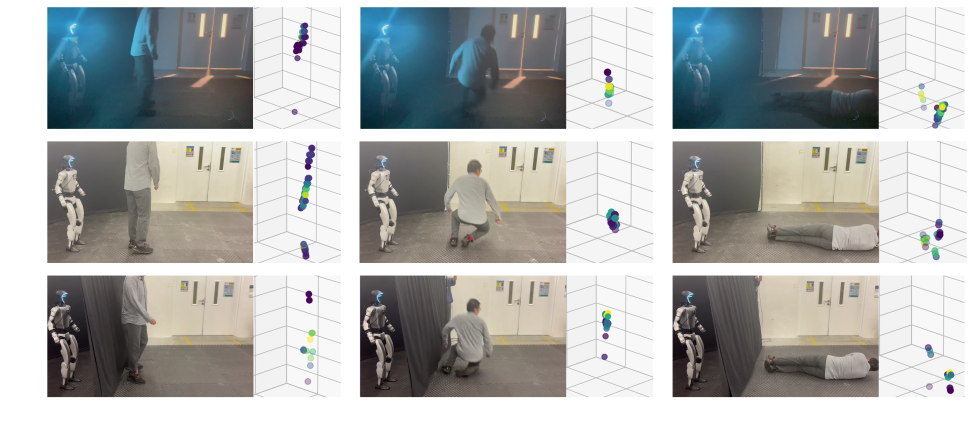}
    \put(2,8.5){(c)}
    \put(2,22){(b)}
    \put(2,35.8){(a)}

    \put(17,0){Pre-fall}
    \put(50.5,0){Drop}
    \put(81,0){Post-fall}
  \end{overpic}
\caption{Visualization of the proposed three-stage fall detection. Columns correspond to the fall phases (pre-fall, drop, post-fall), while rows represent different scenarios: (a) low visibility, (b) normal lighting, and (c) occlusion. Each pair shows the RGB frame (left) and the corresponding mmWave radar point cloud (right).}
  \label{fig4}
\end{figure*}
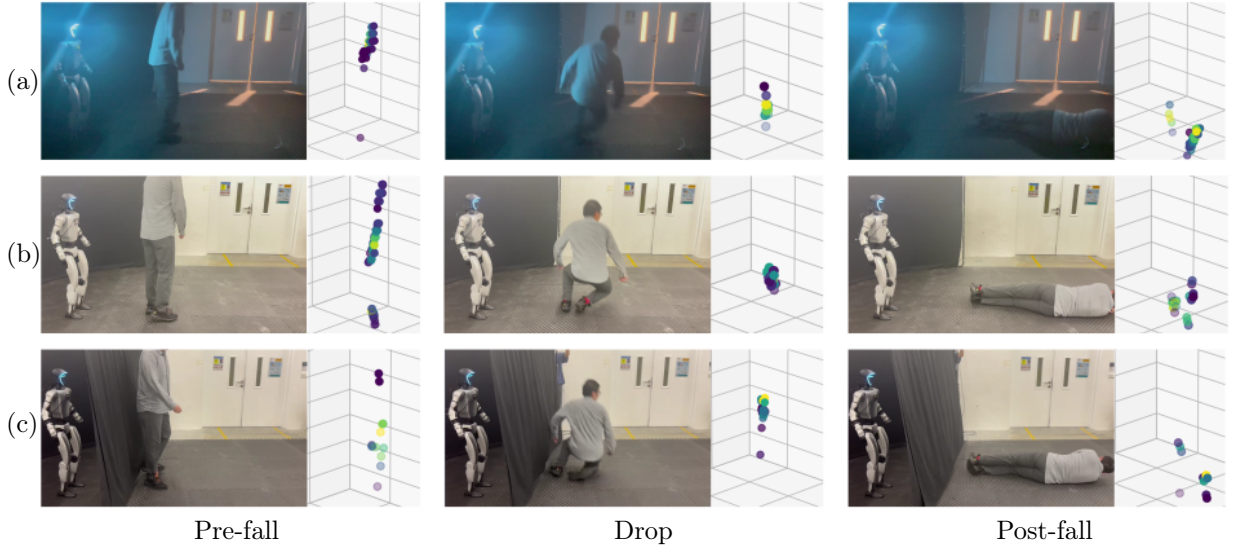

\subsection{Decision and Action Integration}
Finally, we implement a decision module that maps mmWave observations and fused target states to high-level action policies. The module enables online zero-shot robot responses using an LLM, Llama 3.1 8B, conditioned on a predefined prompt template. When a fall event is detected, the module triggers an alert. When the target remains mobile, the robot performs active following by combining the fused spatial state with control information from the people-as-planner module, including the fused target state, the target-to-robot relative distance and relative velocity, and motion planning outputs such as the commanded velocity and subgoal position. Together, these components form an integrated perception--decision--action pipeline that supports long-term companionship and event-driven intervention with improved robustness.

\section{Experiments}

In this section, we systematically evaluate the proposed framework. We assess fall detection performance under diverse environmental conditions and compare against multiple baselines. We further examine the decision module’s ability to produce appropriate responses under different target states. In addition, we conduct an ablation study to quantify the contribution of the temporal modeling module to fall detection performance.

\subsection{Experimental Setup}
Our mobile robotic platform is built on the Unitree G1 robot. The mmWave sensing module uses the TI IWR6843-ISK radar. For visual comparison, we employ an Intel RealSense D455 RGB-D camera. For the LiDAR modality, we adopt the Unitree L1 LiDAR.
\begin{table}[t]
\centering
\setlength{\tabcolsep}{6pt}
\renewcommand{\arraystretch}{1.15}
\caption{Cross-modality availability across eight indoor scenes. 
We report whether reliable fall detection can be achieved by each method in scenes S1--S8 (defined in Fig.~\ref{fig2}).}
\resizebox{\columnwidth}{!}{%
\begin{tabular}{lcccccccc}
\toprule
\textbf{Method} &
\textbf{S1} & \textbf{S2} & \textbf{S3} & \textbf{S4} &
\textbf{S5} & \textbf{S6} & \textbf{S7} & \textbf{S8} \\
\midrule
\textbf{YOLO-Pose} & $\checkmark$ & $\times$ & $\times$     & $\checkmark$     & $\times$ & $\times$     & $\checkmark$     & $\times$ \\
\textbf{DR-SPAAM}              & $\checkmark$ & $\checkmark$ & $\times$ & $\checkmark$ & $\checkmark$     & $\times$ & $\checkmark$     & $\checkmark$ \\
\midrule
\textbf{Ours}             & $\checkmark$ & $\checkmark$ & $\checkmark$ & $\checkmark$ & $\checkmark$ & $\checkmark$ & $\checkmark$ & $\checkmark$ \\
\bottomrule
\end{tabular}
}
\label{tab:modality_8scenes}
\end{table}

\begin{table}[t]
\centering
\setlength{\tabcolsep}{2pt}
\renewcommand{\arraystretch}{1.05}
\caption{Evaluation of LLM-based decision making for fall and move cases using structured perception inputs.}
\resizebox{\columnwidth}{!}{%
\begin{tabular}{l|c|c|c}
\toprule
\textbf{Case} & \textbf{Decision Accuracy} & \textbf{Similarity} & \textbf{Numerical Accuracy} \\
\midrule
Fall & 100.0\% & 81.5\% & 100.0\% \\
Move & 100.0\% & 71.5\% & 99.0\% \\
\bottomrule
\end{tabular}
}
\label{tab:llm_zero_shot_capacity}
\end{table}

To evaluate the cross-room generalization of the proposed fall detection framework, we propose the EM-Fall-Home dataset, collected from four volunteers across three scenario types: Normal Lighting, Occlusion, and Low Visibility, across eight distinct indoor environments. For each scenario type, we recorded 30 fall trials and 30 non-fall trials. Non-fall activities include jumping, walking, lying, squatting, and sitting, resulting in 180 motion sequences in total. All sequences were manually annotated with fall and non-fall labels. 
Prior to fall inference, all sequences were processed by the ghost-target and non-human filtering modules. The dataset intentionally includes fall-like activities, particularly lying, squatting, and sitting, which are commonly confused with falls. Each motion sequence lasts 3 seconds, and the six evaluated poses are illustrated in Fig.~\ref{fig5}. 

To train the proposed model and the compared baselines, point cloud frames are converted into feature vector sequences using the descriptors defined in Section III-C. All learnable components are randomly initialized. The models are implemented in PyTorch and optimized using Adam with a constant learning rate of $1\times 10^{-3}$. Training is conducted on an NVIDIA RTX 4070 GPU with a batch size of 32 for 20 epochs.

To improve the reliability of the reported results, we perform 30 random splits for each scenario type. In each split, 20 fall sequences and 20 non-fall sequences are used for training, 5 fall sequences and 5 non-fall sequences are used for validation, and 5 fall sequences and 5 non-fall sequences are used for testing. All evaluation metrics reported below are averaged across the 30 repeated splits.

\begin{table*}[t]
\centering
\renewcommand{\arraystretch}{1.15}
\setlength{\tabcolsep}{5pt}
\caption{Performance comparison of fall detection methods under three environmental scenarios: Normal Lighting, Low Visibility, and Occlusion.}
\begin{tabularx}{\textwidth}{@{} l l *{7}{>{\centering\arraybackslash}X} @{}}
\toprule
\textbf{Scenarios} & \textbf{Method} &
\textbf{Accuracy} & \textbf{TPR} & \textbf{FPR} & \textbf{F1-Score} &
\textbf{AUROC} & \textbf{AUPRC} & \textbf{MCC} \\
\midrule

\multirow{7}{*}{\makecell[l]{Normal\\Lighting}} 
& 4D Image\citep{b9}      & 95.0\% & 96.7\% & 6.7\% & 95.1\% & 99.0\% & 99.0\% & 90.1\% \\
& Elder-Fall\citep{b11}  & 86.0\% & 99.3\% & 27.3\% & 87.6\% & 83.6\% & 73.2\% & 74.7\% \\
& LT-Fall\citep{b12}     & 97.3\% & 94.7\% & \textbf{0.0}\% & 97.3\% & 97.3\% & 97.3\% & 94.8\% \\
& Bi-LSTM\citep{r1}        & 85.7\% & 96.0\% & 24.7\% & 87.0\% & 96.8\% & 97.0\% & 72.9\% \\
& ST-GCN\citep{r2} & 98.0\% & 98.7\% & 2.7\% & 98.0\% & 99.8\% & 99.8\% & 96.0\% \\
& \textbf{Ours} & \textbf{100.0\%} & \textbf{100.0\%} & \textbf{0.0}\% & \textbf{100.0\%} & \textbf{100.0\%} & \textbf{100.0\%} & \textbf{100.0\%} \\
\midrule

\multirow{7}{*}{\makecell[l]{Low\\Visibility}}
& 4D Image\citep{b9}      & 96.3\% & \textbf{100.0\%} & 7.3\% & 96.5\% & \textbf{99.5\%} & \textbf{99.3\%} & 92.9\% \\
& Elder-Fall\citep{b11}  & 82.7\% & 97.3\% & 32.0\% & 84.9\% & 80.7\% & 71.6\% & 68.3\% \\
& LT-Fall\citep{b12}     & 93.7\% & 98.0\% & 10.7\% & 93.9\% & 93.7\% & 89.4\% & 87.7\% \\
& Bi-LSTM\citep{r1}       & 91.7\% & 94.0\% & 10.7\% & 91.9\% & 91.8\% & 92.2\% & 83.4\% \\
& ST-GCN\citep{r2} & 90.0\% & 92.0\% & 12.0\% & 90.2\% & 95.0\% & 93.9\% & 80.1\% \\
& \textbf{Ours} & \textbf{97.0\%} & \textbf{100.0\%} & 6.0\% & \textbf{97.1\%} & 97.0\% & 94.3\% & \textbf{94.2\%} \\
\midrule

\multirow{7}{*}{Occlusion}
& 4D Image\citep{b9}      & 94.0\% & 96.7\% & 8.7\% & 94.2\% & 99.1\% & \textbf{99.1\%} & 88.1\% \\
& Elder-Fall\citep{b11}  & 72.9\% & 90.7\% & 46.7\% &77.7\% & 70.7\% & 67.7\% & 47.8\% \\
& LT-Fall\citep{b12}     & 86.0\% & 86.0\% & 14.0\% & 86.0\% & 86.0\% & 81.0\% & 72.0\% \\
& Bi-LSTM\citep{r1}        & 84.7\% & 99.3\% & 30.0\% & 86.6\% & 96.2\% & 94.0\% & 72.5\% \\
& ST-GCN\citep{r2} & 94.3\% & 92.7\% & \textbf{4.0\%} & 94.2\% & 98.6\% & 98.9\% & 88.7\% \\
& \textbf{Ours} & \textbf{94.7\%} & \textbf{100.0\%} & 10.7\% & \textbf{94.9\%} & 94.7\% & 90.4\% & \textbf{89.8\%} \\
\bottomrule
\end{tabularx}
\label{tab:scenario_metrics}
\end{table*}

\subsection{Fall Detection Performance Comparison}
To evaluate the effectiveness of the proposed fall detection method, we compared our method with other representative mmWave-based baselines: 4D Image\citep{b9}, Elder-Fall~\citep{b11}, LT-Fall~\citep{b12}, Bi-LSTM\citep{r1}, and ST-GCN\citep{r2}. We set $\tau_{\Delta h}=-0.5$, $\tau_{w}=0.4$, and $\tau_{r}=0.6$ for all scenarios. Table~\ref{tab:scenario_metrics} summarizes the quantitative comparison between our method and the baselines under three environmental conditions: normal lighting, low visibility, and occlusion. First, our method achieves a fall detection TPR of 100\% in all three settings, resulting in zero missed detections. This property is particularly important for fall detection systems, where false negatives incur a high safety risk.
Second, while maintaining perfect TPR, our method achieves the highest F1-score, which is 97.1\% and 94.9\%, and MCC, which is 94.2\% and 89.8\%, in the two challenging scenarios of low visibility and occlusion. These results indicate that the improvements are not obtained at the cost of precision, but reflect better suppression of false alarms and more consistent decision behavior.
\begin{table}[htp]
\centering
\caption{Ablation study of the temporal fall analysis module.}
\label{tab:ablation_temporal_analysis}
\resizebox{\columnwidth}{!}{%
\begin{tabular}{lccc}
\toprule
\textbf{Setting} & \textbf{F1} & \textbf{AUPRC} & \textbf{MCC} \\
\midrule
w/o temporal analysis      & 96.6\% & 93.4\% & 93.1\% \\
w/o post-fall              & 96.9\% & 93.9\% & 93.8\% \\
w/o pre-fall               & \textbf{97.0\%} & \textbf{94.1\%} & \textbf{94.0\%} \\
\textbf{Complete temporal analysis} & 97.3\% & 94.7\% & 94.6\% \\
\bottomrule
\end{tabular}
}
\end{table}

\begin{table}[htp]
\centering
\caption{Ablation study on the depth of the MLP classifier.}
\label{tab:ablation_mlp_depth}
\resizebox{\columnwidth}{!}{%
\begin{tabular}{lcccc}
\toprule
\textbf{Setting} & \textbf{F1-Score} & \textbf{AUROC} & \textbf{AUPRC} & \textbf{MCC} \\
\midrule
2 layers & 90.7\% & 89.9\% & 83.4\% & 81.1\% \\
\textbf{3 layers} & \textbf{97.3\%} & \textbf{97.2\%} & \textbf{94.7\%} & \textbf{94.6\%} \\
4 layers & 96.2\% & 96.1\% & 92.9\% & 92.4\% \\
\bottomrule
\end{tabular}
}
\end{table}

\begin{table}[htp]
\centering
\caption{Ablation study on the temporal window duration used for fall analysis.}
\label{tab:ablation_temporal_duration}
\resizebox{\columnwidth}{!}{%
\begin{tabular}{lcccc}
\toprule
\textbf{Setting} & \textbf{F1-Score} & \textbf{AUROC} & \textbf{AUPRC} & \textbf{MCC} \\
\midrule
0.5 s & 95.9\% & 95.8\% & 92.2\% & 91.9\% \\
\textbf{1.0 s} & \textbf{97.3\%} & \textbf{97.2\%} & \textbf{94.7\%} & \textbf{94.6\%} \\
1.5 s & 95.0\% & 94.9\% & 91.4\% & 90.0\% \\
\bottomrule
\end{tabular}
}
\end{table}
Moreover, the performance degradation of our approach in complex scenes remains moderate, whereas several baselines exhibit either a sharp increase in FPR under occlusion or a noticeable drop in TPR. Our method maintains the best overall metrics with a relatively low false positive rate, demonstrating stronger robustness under observation degradation.
Overall, these results suggest that our method is well-suited for cross-area and cross-environment deployment in real homes and mobile robotic platforms. It maintains reliable fall detection under environmental changes, prioritizes avoiding missed falls, and keeps false alarms within an acceptable range. Visualization examples are shown in Fig.~\ref{fig4}.

Finally, to highlight the advantage of mmWave sensing under challenging visible conditions, we further compare our method with representative baselines from other sensing modalities. Specifically, we consider a vision baseline based on YOLO-Pose~\citep{b50} with a temporal fall detection head, and a LiDAR baseline based on DR-SPAAM~\citep{b51} using the same temporal fall detection scheme. All methods are evaluated under eight different visibility scenarios, and the results are summarized in Table~\ref{tab:modality_8scenes}.

\subsection{Decision Capacity Experiment}
To evaluate the decision generation capability of the proposed decision module, we select 200 structured perception instances, including 100 fall events and 100 normal motion states captured from the dataset. These instances are provided to the module, and we quantitatively measure the correctness of the generated decisions. Each input is represented in a structured format and includes the fused target state, the target--robot relative distance and velocity, and the motion planner outputs (commanded velocity and subgoal position). The results are reported in Table~\ref{tab:llm_zero_shot_capacity}. Decision Accuracy measures correct state recognition; Similarity denotes the cosine similarity between generated responses and reference answers; Numerical Accuracy evaluates the correctness of extracted numerical information.

The decision module achieves a 100\% success rate in both the Fall and Move scenarios, indicating reliable correctness in state recognition and numerical information extraction. Meanwhile, the Similarity score in the Fall scenario is 81.5\%, whereas it is 71.5\% in the Move scenario. This difference arises because fall-related responses follow a more consistent alert policy, whereas normal motion responses allow greater linguistic variability. As a result, the generated outputs remain semantically correct but exhibit larger textual variation, leading to lower similarity scores. Moreover, because the subgoal position was missing for one Move instance, the Numerical Accuracy decreased slightly, which is 99.0\%.

\subsection{Ablation Experiments}
To quantify the contribution of each component in the proposed three-stage fall detection scheme, we conduct an ablation study including (i) the temporal analysis stages, (ii) the MLP depth, and (iii) the temporal window duration.
When aggregating all scenarios, Table~\ref{tab:ablation_temporal_analysis} shows that the full three-stage temporal decision provides a clear performance gain. The complete model attains an F1-score of 97.3\%, an AUPRC of 94.7\%, and a MCC of 94.6\%. Removing either the post-fall or pre-fall analysis consistently degrades performance, indicating that relying on instantaneous cues alone weakens the ability to distinguish true falls from fall-like actions. The impact of removing post-fall is slightly more pronounced, suggesting that post-impact pose stability and spatial structure cues play an important role in suppressing false positives.
Table~\ref{tab:ablation_mlp_depth} reports the MLP-depth ablation, where a 3-layer MLP achieves the best performance. The 2-layer variant underfits, with an F1-score of 90.7\%, while increasing the depth to 4 layers leads to a slight performance drop, with an F1-score of 96.2\%, suggesting reduced generalization due to redundant parameters.
Table~\ref{tab:ablation_temporal_duration} evaluates the temporal window duration. A 1.0 s window performs best. A 0.5 s window provides insufficient temporal context, leading to an F1-score of 95.9\%, whereas a 1.5 s window tends to include additional non-fall motion fragments that degrade fall detection performance, resulting in an F1-score of 95.0\% and an MCC of 90.0\%.

\section{Conclusion}

This work presents EM-Fall, an embodied mobile mmWave-based fall detection framework for in-home elderly care. By deploying radar sensing on a humanoid robotic platform, the system overcomes the coverage gaps and viewpoint limitations of static radar installations, enabling continuous and privacy-preserving monitoring in real residential environments.
The framework integrates spatial state estimation, human-centered perception and fall inference, and decision integration within a unified robotic pipeline. mmWave target tracking is fused with robot ego-motion to maintain consistent target states during mobile operation, while human-centered perception suppresses non-human entities and multipath artifacts and temporal inference distinguishes true falls from fall-like activities.
Experiments across eight indoor environments with four participants under normal lighting, occlusion, and low-visibility conditions demonstrate robust fall detection performance and stable observation under diverse environmental conditions. These results suggest that embodied mobile mmWave sensing provides a practical perception–detection–decision foundation for long-term home companion robots.

\clearpage
\newpage
\bibliographystyle{assets/plainnat}
\bibliography{paper}

@article{b1,
  title={Global, regional, and national burden of falls among older adults: findings from the Global Burden of Disease Study 2021 and Projections to 2040},
  author={Chen, Yang and Dai, Feifei and Huang, Shulun and Qi, Daoda and Peng, Chengyi and Zhang, Aijia and Wang, Yuan and Gu, Yan and Guo, Jingjing},
  journal={npj Aging},
  volume={11},
  number={1},
  pages={85},
  year={2025},
  publisher={Nature Publishing Group UK London}
}

@article{b33,
  title={Vision-based human fall detection systems: A review},
  author={Benkaci, Asma and Sliman, Layth and Dellys, Hachemi Nabil},
  journal={Procedia Computer Science},
  volume={241},
  pages={203--211},
  year={2024},
  publisher={Elsevier}
}

@article{b34,
  title={Survey on context-aware radio frequency-based sensing},
  author={Casmin, Eugene and Oliveira, Rodolfo},
  journal={Sensors},
  volume={25},
  number={3},
  pages={602},
  year={2025},
  publisher={MDPI}
}

@article{b17,
  title={Radar-based fall detection: A survey [survey]},
  author={Hu, Shuting and Cao, Siyang and Toosizadeh, Nima and Barton, Jennifer and Hector, Melvin G and Fain, Mindy J},
  journal={IEEE robotics \& automation magazine},
  volume={31},
  number={3},
  pages={170--185},
  year={2024},
  publisher={IEEE}
}

@article{b23,
  title={Fall detection system using millimeter-wave radar based on neural network and information fusion},
  author={Yao, Yicheng and Liu, Changyu and Zhang, Hao and Yan, Baiju and Jian, Pu and Wang, Peng and Du, Lidong and Chen, Xianxiang and Han, Baoshi and Fang, Zhen},
  journal={IEEE Internet of Things Journal},
  volume={9},
  number={21},
  pages={21038--21050},
  year={2022},
  publisher={IEEE}
}

@article{b9,
  title={Non-contact fall detection system using 4D imaging radar for elderly safety based on a CNN model},
  author={Ahn, Sejong and Choi, Museong and Lee, Jongjin and Kim, Jinseok and Chung, Sungtaek},
  journal={Sensors},
  volume={25},
  number={11},
  pages={3452},
  year={2025},
  publisher={MDPI}
}

@article{b11,
  title={Millimeter-wave radar-based elderly fall detection fed by one-dimensional point cloud and Doppler},
  author={Kittiyanpunya, Chainarong and Chomdee, Pongsathorn and Boonpoonga, Akkarat and Torrungrueng, Danai},
  journal={Ieee Access},
  volume={11},
  pages={76269--76283},
  year={2023},
  publisher={IEEE}
}

@article{b12,
  title={Lt-fall: The design and implementation of a life-threatening fall detection and alarming system},
  author={Zhang, Duo and Zhang, Xusheng and Li, Shengjie and Xie, Yaxiong and Li, Yang and Wang, Xuanzhi and Zhang, Daqing},
  journal={Proceedings of the ACM on Interactive, Mobile, Wearable and Ubiquitous Technologies},
  volume={7},
  number={1},
  pages={1--24},
  year={2023},
  publisher={ACM New York, NY, USA}
}

@article{b45,
  title={FMCW radar point cloud multiperson tracking using a Kalman filter--based approach},
  author={Santos, Bruno and Oliveira, AS and Carvalho, Nuno Borges and Fernandes, Rui and Cannizzaro, Andrea and Cruz, Pedro Miguel},
  journal={URSI Radio Sci. Lett.},
  volume={3},
  pages={68},
  year={2021}
}

@article{b46,
  title={Following is all you need: Robot crowd navigation using people as planners},
  author={Liao, Yuwen and Xu, Xinhang and Bai, Ruofei and Yang, Yizhuo and Cao, Muqing and Yuan, Shenghai and Xie, Lihua},
  journal={IEEE Robotics and Automation Letters},
  year={2025},
  publisher={IEEE}
}

@article{b24,
  title={Multi-object tracking with mmwave radar: A review},
  author={Pearce, Andre and Zhang, J Andrew and Xu, Richard and Wu, Kai},
  journal={Electronics},
  volume={12},
  number={2},
  pages={308},
  year={2023},
  publisher={MDPI}
}

@article{b28,
  title={Split BiRNN for real-time activity recognition using radar and deep learning},
  author={Werthen-Brabants, Lorin and Bhavanasi, Geethika and Couckuyt, Ivo and Dhaene, Tom and Deschrijver, Dirk},
  journal={Scientific Reports},
  volume={12},
  number={1},
  pages={7436},
  year={2022},
  publisher={Nature Publishing Group UK London}
}

@inproceedings{b50,
  title={Yolo-pose: Enhancing yolo for multi person pose estimation using object keypoint similarity loss},
  author={Maji, Debapriya and Nagori, Soyeb and Mathew, Manu and Poddar, Deepak},
  booktitle={Proceedings of the IEEE/CVF conference on computer vision and pattern recognition},
  pages={2637--2646},
  year={2022}
}

@inproceedings{b51,
  title={DR-SPAAM: A spatial-attention and auto-regressive model for person detection in 2D range data},
  author={Jia, Dan and Hermans, Alexander and Leibe, Bastian},
  booktitle={2020 IEEE/RSJ International Conference on Intelligent Robots and Systems (IROS)},
  pages={10270--10277},
  year={2020},
  organization={IEEE}
}

@article{x1,
  title={Millimeter-wave radar monitoring for elder’s fall based on multi-view parameter fusion estimation and recognition},
  author={Feng, Xiang and Shan, Zhengliang and Zhao, Zhanfeng and Xu, Zirui and Zhang, Tianpeng and Zhou, Zihe and Deng, Bo and Guan, Zirui},
  journal={Remote Sensing},
  volume={15},
  number={8},
  pages={2101},
  year={2023},
  publisher={MDPI}
}

@article{x2,
  title={Leveraging falling acceleration and body part clustering for physics-based human fall detection with millimeter wave radar},
  author={Huh, Hyunsuk and Jeong, Iljoo and Lee, Anna and Lee, Seungchul and Shin, Young-Sik},
  journal={Engineering Applications of Artificial Intelligence},
  volume={159},
  pages={111500},
  year={2025},
  publisher={Elsevier}
}

@inproceedings{c18,
  title={A lightweight neural-net with assistive mobile robot for human fall detection system},
  author={Chin, Wei Hong and Tay, Noel Nuo Wi and Kubota, Naoyuki and Loo, Chu Kiong},
  booktitle={2020 International Joint Conference on Neural Networks (IJCNN)},
  pages={1--6},
  year={2020},
  organization={IEEE}
}

@article{c20,
  title={New eldercare robot with path-planning and fall-detection capabilities},
  author={Elwaly, Ahmad and Abdellatif, A and El-Shaer, Y},
  journal={Applied Sciences},
  volume={14},
  number={6},
  pages={2374},
  year={2024},
  publisher={MDPI}
}

@article{r1,
  title={Continuous human activity classification from FMCW radar with Bi-LSTM networks},
  author={Shrestha, Aman and Li, Haobo and Le Kernec, Julien and Fioranelli, Francesco},
  journal={IEEE Sensors Journal},
  volume={20},
  number={22},
  pages={13607--13619},
  year={2020},
  publisher={IEEE}
}

@inproceedings{r2,
  title={Spatial temporal graph convolutional networks for skeleton-based action recognition},
  author={Yan, Sijie and Xiong, Yuanjun and Lin, Dahua},
  booktitle={Proceedings of the AAAI conference on artificial intelligence},
  volume={32},
  number={1},
  year={2018}
}

@article{r3,
  title={Semisupervised human activity recognition with radar micro-Doppler signatures},
  author={Li, Xinyu and He, Yuan and Fioranelli, Francesco and Jing, Xiaojun},
  journal={IEEE Transactions on Geoscience and Remote Sensing},
  volume={60},
  pages={1--12},
  year={2021},
  publisher={IEEE}
}

@article{b30,
  title={The methods of fall detection: A literature review},
  author={Newaz, Nishat Tasnim and Hanada, Eisuke},
  journal={Sensors},
  volume={23},
  number={11},
  pages={5212},
  year={2023},
  publisher={MDPI}
}

@article{b32,
  title={Monitoring wearable devices for elderly people with dementia: a review},
  author={Rocha, In{\^e}s C and Arantes, Marcelo and Moreira, Ant{\'o}nio and Vila{\c{c}}a, Jo{\~a}o L and Morais, Pedro and Matos, Dem{\'e}trio and Carvalho, V{\'\i}tor},
  journal={Designs},
  volume={8},
  number={4},
  pages={75},
  year={2024},
  publisher={MDPI}
}

\end{document}